\documentclass[letterpaper]{article} 
\usepackage{aaai2026}  
\usepackage{times}  
\usepackage{helvet}  
\usepackage{courier}  
\usepackage[hyphens]{url}  
\usepackage{graphicx} 
\usepackage{natbib}  
\usepackage{caption} 
\usepackage{algorithm}
\usepackage{algorithmic}

\usepackage{amsfonts}
\usepackage{amsmath}
\usepackage{booktabs}
\usepackage{makecell}
\usepackage{multirow}

\usepackage{tcolorbox}
\definecolor{takeawaycol}{HTML}{2D6E6A}  
\newtcolorbox{takeaway}{
  colback=takeawaycol!10, colframe=takeawaycol,
  leftrule=0pt, rightrule=0pt, toprule=0pt, bottomrule=0pt,
  boxsep=3pt, left=5pt, right=5pt, top=3pt, bottom=3pt,
}

\usepackage{newfloat}
\usepackage{listings}
\DeclareCaptionStyle{ruled}{labelfont=normalfont,labelsep=colon,strut=off} 
\floatstyle{ruled}
\newfloat{listing}{tb}{lst}{}
\floatname{listing}{Listing}
\nocopyright 

\title{Revisiting Tree Search for LLMs: Gumbel and Sequential Halving for Budget-Scalable Reasoning}
\author{
    Leonid Ugadiarov\textsuperscript{\rm 1,2},
    Yuri Kuratov\textsuperscript{\rm 1,2},
    Aleksandr Panov\textsuperscript{\rm 1,2},
    Alexey Skrynnik\textsuperscript{\rm 1,2}
}
\affiliations{
    \textsuperscript{\rm 1}AXXX, Moscow, Russia\\
    \textsuperscript{\rm 2}MIRAI, Moscow, Russia
}

\begin{document}

\maketitle

\begin{abstract}
Neural tree search is a powerful decision-making algorithm widely used in complex domains such as game playing and model-based reinforcement learning. Recent work has applied AlphaZero-style tree search to enhance the reasoning capabilities of Large Language Models (LLMs) during inference, but we find that this approach suffers from a scaling failure: on GSM8K and Game24, accuracy drops as the search budget increases. In this paper, we present ReSCALE, an adaptation of Gumbel AlphaZero MCTS that replaces Dirichlet noise and PUCT selection with Gumbel sampling and Sequential Halving, restoring monotonic scaling without changes to the model or its training. ReSCALE reaches 58.4\% on GSM8K and 85.3\% on Game24 at budgets where the baseline degrades. Ablations confirm that Sequential Halving is the primary driver of the improvement.
\end{abstract}

 \begin{links}
     \small
     \link{Code}{https://github.com/CognitiveAISystems/ReSCALE}
 \end{links}

\section{Introduction}

Large language models (LLMs) have achieved remarkable results on reasoning benchmarks~\citep{openai2024openai,guo2025deepseek}. However, even such reasoning-oriented models still struggle with problems that require multi-step reasoning and planning~\citep{wan2024logicasker,kambhampati2024position,valmeekam2024llms}. A common way to enable step-by-step reasoning in non-reasoning LLMs is chain-of-thought (CoT) prompting~\citep{wei2022chain}, where the model is asked to generate intermediate reasoning steps before providing an answer. CoT can substantially improve non-reasoning LLMs' accuracy on arithmetic, logical, and general tasks, but in its standard form it still represents a single reasoning attempt: the model follows one sampled trajectory of thoughts, with no mechanism to explore alternatives and backtrack.

\begin{figure}[ht!]
    \centering
    \includegraphics[width=1.0\columnwidth]{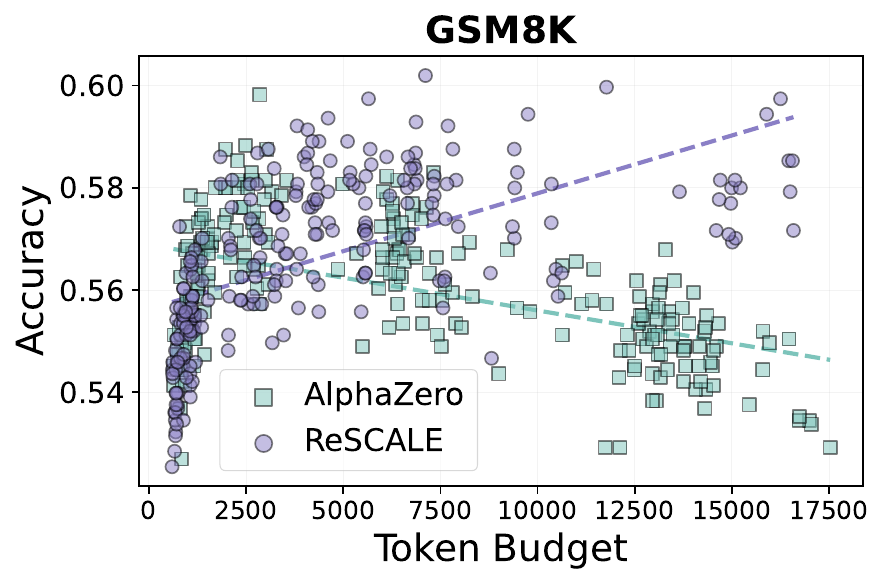}
    \vspace{-10px}
    \caption{\textbf{Gumbel + Sequential Halving enables scaling of MCTS for reasoning.} Comparison of tree-search methods for LLM reasoning on GSM8K across increasing token budgets. The standard AlphaZero-style approach plateaus and declines at higher budgets, while proposed ReSCALE decoding achieves sustained accuracy gains, demonstrating better scaling with additional compute.}\label{fig:gsm8k_scaling}
    \vspace{-10px}
\end{figure}

To move beyond this one-shot CoT setting, recent work augments LLMs with search over multiple reasoning trajectories, ranging from sampling multiple CoT traces~\citep{wang2023selfconsistency} to more structured tree-search LLM methods~\citep{yao2023tree,hao2023reasoning}. In these methods, models explore alternative reasoning paths before producing a final answer.
Tree-search-based approaches are appealing because they provide a principled way to allocate test-time
  compute~\citep{snell2025scaling}, trading off exploration and exploitation under a limited
  budget~\citep{pallagani2024prospects}.

This idea builds on a long line of progress in Monte Carlo Tree Search (MCTS), from the upper confidence bound for trees algorithm~\citep{kocsis2006bandit} to AlphaZero-style methods that incorporate learned policies and value functions to achieve superhuman play in Go, chess, shogi, and Atari~\citep{silver2017mastering,schrittwieser2020mastering}, and even to the discovery of improved matrix-multiplication algorithms~\citep{fawzi2022discovering}.

\citet{wan2024alphazero} apply AlphaZero-like tree search to LLM decoding, using a learned value model to guide MCTS. With small to moderate simulation budgets, ranging from 5 to 15 simulations and involving around 500 tokens, this approach substantially outperforms standard decoding strategies. However, we show that when the computation budget is scaled beyond this range by increasing the number of simulations or the search depth, accuracy on the GSM8K dataset plateaus and even slightly declines (Figure~\ref{fig:gsm8k_scaling}), while the number of visited nodes and generated tokens continues to grow. We observe similar degradation at large budgets on the Game24 dataset. This behavior resembles the lookahead pathology phenomenon, where increased search effort can paradoxically degrade decision quality~\citep{nguyen2024lookahead}.

A well-designed search algorithm should exhibit the opposite behavior: performance should improve as the search budget increases. Classical work on planning and MCTS has introduced techniques that are well suited to this fixed-budget perspective, including Gumbel-based root sampling~\citep{danihelka2022policy} and Sequential Halving~\citep{karnin2013almost} algorithms for best-arm identification. To the best of our knowledge, Gumbel sampling and Sequential Halving have not yet been integrated into AlphaZero-like tree-search frameworks for LLM decoding. In this study, we examine the effects of incorporating these two techniques into such a framework. 
We find that these components can significantly improve performance with larger simulation budgets.

\textbf{Our contributions are:}

\begin{itemize}
  \item We identify a scaling failure in AlphaZero-style tree search for LLM reasoning: on GSM8K and Game24, accuracy \emph{decreases} beyond a moderate simulation budget, even as the algorithm explores more tree nodes.
  \item We trace this failure to the action selection mechanism and show that replacing it with Gumbel-based sampling and Sequential Halving restores monotonic scaling, without any changes to the model or its training.
  \item We extensively evaluate ReSCALE on GSM8K and Game24, demonstrating that it effectively converts additional compute into accuracy gains in regimes where the baseline degrades.
\end{itemize}

\section{Method}
\begin{figure*}[ht!]
\centering
\includegraphics[width=1.0\textwidth]{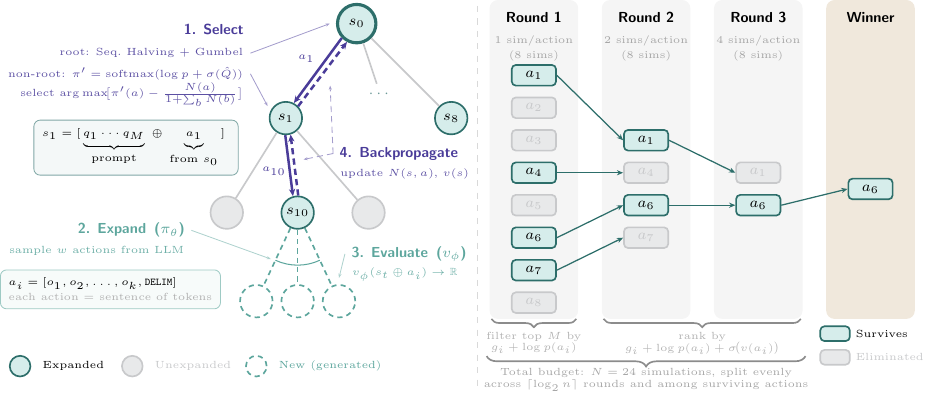}
\caption{ \textbf{Left:} A single simulation in tree search with sentence-level actions. Each simulation traverses from the root to a leaf, selecting actions via Sequential Halving with Gumbel noise at the root and an improved policy at non-root nodes. The selected leaf is expanded by sampling $w$ actions from the LLM ($\pi_\theta$), each evaluated by the value network $v_{\phi}$, and the resulting values are backpropagated to update ancestor statistics.
 \textbf{Right:} Sequential Halving at the root node. From $M = 8$ candidate actions, the budget of $N = 24$ simulations is split evenly across $\lceil\log_2 M\rceil = 3$ rounds. After each round, the bottom half of actions (by score $g_i + \log p(a_i) + \sigma(v(a_i))$) is eliminated, until a single winner remains.}\label{fig:gumbel_mcts}
 \vspace{-1em}
\end{figure*}

We first describe the background components shared with prior work: the MDP formulation, action space, value network, and standard AlphaZero tree search, and then present our Gumbel AlphaZero adaptation for LLM decoding.

\subsection{Background}

\paragraph{MDP Formulation}
We formulate language generation as a token-level Markov Decision Process (MDP)~\citep{puterman1994markov}.
Both actions and states reside in token space: the state $s_t = [q_1, \dots, q_M, o_0, \dots, o_{t-1}]$ concatenates the input prompt and previously generated tokens, and the LLM serves as a policy $\pi_{\theta}^{token}$ that autoregressively selects the next token $o_t$ from vocabulary $\mathcal{O}$, yielding a deterministic transition $s_{t+1} = s_t \oplus o_t$.

The reward function $r(s_t)$ represents the task objective (defined by rules or a pre-trained reward model) and can be sparse (e.g., binary correctness of the final state $s_T$) or dense.
Our method uses tree search to guide LLM generation toward maximizing the cumulative reward $\sum_{t=0}^{T} r(s_t)$.

\paragraph{Action Space}
Following \citet{wan2024alphazero}, we define an action as a sequence of tokens ending with a special delimiter \texttt{DELIM} (e.g., a newline) or a \texttt{STOP} token. The full model output is thus a sequence of sentence-level actions $[a_0, a_1, \dots, a_{d-1}]$.
Compared to token-level actions ($a_i = o_i$), this reduces the search tree depth at the cost of a larger branching factor. During tree expansion, we sample $w$ actions and normalize their probabilities to sum to 1, denoting the result as $p(a)$. We also restrict the maximum tree depth to $d$, making the search computationally tractable.

\paragraph{Value Network} AlphaZero-like tree search~\citep{silver2018general} requires a value network $v_{\phi}$ that estimates the expected cumulative reward from a given state. We use a value network that shares the policy architecture but adds a scalar value head. The network is trained with Monte Carlo return targets~\citep{sutton2018reinforcement} using a mean-squared error objective; training details are provided in Section~\ref{sec:experiments}.

\paragraph{AlphaZero Tree Search}
The standard AlphaZero tree search~\citep{silver2018general} performs multiple simulations from the root state $s_t$, where each simulation is a single traversal from the root to an unexpanded node. During traversal, actions are selected via the PUCT algorithm~\citep{rosin2011multi}; at the unexpanded node, new children are added to the tree and the value estimate is backpropagated to all ancestors.
After all simulations, the action is chosen proportionally to exponentiated visit counts: $a_{t+1} \sim N(s_t, a)^{1/\tau} / \sum_b N(s_t, b)^{1/\tau}$, where $N(s_t, a)$ is the number of times action $a$ was selected from $s_t$.

\subsection{ReSCALE: Gumbel MCTS with Sequential Halving for LLM Decoding}

An overview of the ReSCALE (Reasoning via Scalable Compute Allocation for LLM Exploration) approach is shown in Figure~\ref{fig:gumbel_mcts}. Compared to the standard AlphaZero tree search, Gumbel AlphaZero~\citep{danihelka2022policy} replaces Dirichlet noise and PUCT-based selection at the root with Gumbel noise and Sequential Halving, uses an improved policy with mixed value approximation at non-root nodes, and selects the final action deterministically. 

\paragraph{Selection at the Root Node}
Each action $a^i$ at the root is scored by $f^i = g(a^i) + \log p(a^i)$, where $g(a^i)$ is sampled Gumbel noise.
After simulations, scores are refined with the value-based monotonic function $\sigma$, where $c_{\text{visit}} = 50$:
\[
    \sigma(v(s_t \oplus a)) = (c_{\text{visit}} + \max_{a^{*}} N(s_t, a^{*})) \, v(s_t \oplus a).
\]

\begin{takeaway}
\textbf{\textit{Takeaway:}} Gumbel noise enables sampling actions without replacement while providing policy improvement guarantees, replacing the heuristic Dirichlet exploration noise of standard AlphaZero.
\end{takeaway}

\noindent The simulation budget $N$ is distributed via Sequential Halving. Starting with the top $M \leq w$ actions ranked by their Gumbel scores, the procedure runs $N / \lceil\log_2 M\rceil$ simulations split evenly among surviving actions, then eliminates the bottom half based on updated scores $f^i + \sigma(v(s_t \oplus a^i))$.
This repeats until a single action remains: $s_{t+1} = s_t \oplus a_{\text{next}}$.

\begin{takeaway}
\textbf{\textit{Takeaway:}} Sequential Halving focuses the simulation budget on the most promising actions by progressively eliminating weaker ones,
   improving efficiency --- important given the cost of LLM forward passes.
\end{takeaway}

\paragraph{Selection at Non-Root Nodes}
At a non-root node $s_t$, actions are selected using an improved policy that combines the prior action probabilities with value estimates:
\[
    \pi'(a) = \mathrm{softmax}\!\bigl(\log p(a) + \sigma(v_c(s_t \oplus a))\bigr),
\]
where $v_c$ is a \emph{completed} value: for visited actions ($N(s_t, a) > 0$) we use the backpropagated value $v(s_t \oplus a)$, and for unvisited actions we use a mixed approximation:
\vspace{-0.2em}
\begin{equation*}
    v_c(s_t \oplus a) = \begin{cases}
      v(s_t \oplus a), & \text{if } N(s_t, a) > 0, \\
      v_{\text{mix}}(s_t \oplus a), & \text{otherwise}.
    \end{cases}
\end{equation*}
Let $N_\Sigma = \sum_b N(s_t, b)$ denote the total visit count at $s_t$ and $\bar{v}_t = \frac{\sum_{b:\,N(s_t,b)>0} p(b)\,v(s_t \oplus b)}{\sum_{b:\,N(s_t,b)>0} p(b)}$ the policy-weighted mean value over visited sibling actions, serving as an empirical baseline for unvisited ones. Then:
\[
    v_{\text{mix}}(s_t \oplus a) = \frac{v_{\phi}(s_t \oplus a) + N_\Sigma \cdot \bar{v}_t}{1 + N_\Sigma}.
\]
The final action is chosen in a deterministic way according to $\pi'$, with a correction that uses the normalized visit counts:
\[
a_{\text{next}} = \arg\max_a \left( \pi'(a) - \frac{N(s_t, a)}{1 + \sum_b N(s_t, b)} \right).
\]

\begin{takeaway}
\textbf{\textit{Takeaway:}} The mixed value approximation interpolates between the value network's prediction (when few actions are visited) and the observed mean of visited siblings (as data accumulates), preventing overcommitment to inaccurate estimates for unexplored actions.   
\end{takeaway}

\paragraph{Expansion and Evaluation}
When the selection phase reaches a non-terminal leaf node $s_t$, we expand it by sampling $w$ actions $\{a_i\}_{i=0}^{w-1}$ from the LLM and adding child nodes $\{s_t \oplus a_i\}_{i=0}^{w-1}$ to the tree.
Each child is then evaluated by the value network, yielding estimates $v_{\phi}(s_t \oplus a_i)$ that are used in backpropagation.

\paragraph{Backpropagation}
Once a leaf is expanded and evaluated, its value $v_{\text{leaf}}$ is propagated upward along the path back to the root. At each ancestor node $s$ on this path, the visit count $N(s, a)$ for the action $a$ that was chosen at $s$ is incremented, and the value estimate is updated as a running mean:
\[
    v(s) \leftarrow \frac{v(s) \cdot N(s, a) + v_{\text{leaf}}}{N(s, a) + 1}.
\]
Here $v_{\text{leaf}}$ is either $v_{\phi}$ for non-terminal leaves or the environment reward for terminal states. The same $v_{\text{leaf}}$ is used at every node along the path.

\begin{table*}[t!]
    \centering
    \begin{tabular*}{\textwidth}{@{\extracolsep{\fill}}llccccccc}
        \toprule
        & & \multicolumn{3}{c}{GSM8K} & \multicolumn{3}{c}{Game24} \\
        \cmidrule(lr){3-5} \cmidrule(lr){6-8}
        Tree Search & Budget                  & Tokens                    & Acc., \%       & Max.\ Acc., \% & Tokens                    & Acc., \%        & Max.\ Acc., \% \\
        \midrule
        AlphaZero   & \multirow{2}{*}{Small}  & \multirow{2}{*}{0.5K--2K} & $56.0 \pm 1.2$ & $58.8$ & \multirow{2}{*}{0.2K--2K} & $74.4 \pm 12.0$ & $86.7$ \\
        ReSCALE     &                         &                           & $55.1 \pm 1.1$ & $58.6$         &                           & $71.6 \pm 10.5$ & $81.2$         \\
        \midrule
        AlphaZero   & \multirow{2}{*}{Medium} & \multirow{2}{*}{6K--8K}   & $56.7 \pm 0.9$ & $58.3$         & \multirow{2}{*}{2K--4K}   & $84.3 \pm 1.0$  & $86.2$ \\
        ReSCALE     &                         &                           & $57.9 \pm 1.0$ & $60.2$         &                           & $83.4 \pm 1.4$  & $85.9$         \\
        \midrule
        AlphaZero   & \multirow{2}{*}{Large}  & \multirow{2}{*}{16K--18K} & $53.6 \pm 0.7$ & $55.0$ & \multirow{2}{*}{4K--6K}   & $82.9 \pm 1.1$  & $84.8$ \\
        ReSCALE     &                         &                           & $58.4 \pm 0.9$ & $59.7$         &                           & $85.3 \pm 0.6$  & $85.9$         \\
        \midrule
        Best-of-N   & N = 32                  & 3.5K                      & $53.4 \pm 0.4$ & -              & 2.5K                      & $54.1 \pm 1.0$  & -              \\
        \bottomrule
    \end{tabular*}
    \caption{Combined results on GSM8K and Game24. For each budget level, mean accuracy $\pm$ std is computed across hyperparameter configurations; for Best-of-N, across three random seeds. Best single-configuration accuracy (Max.\ Acc.) is also reported. On GSM8K, AlphaZero accuracy drops at Large budget (53.6\%) while ReSCALE continues to improve (58.4\%), both surpassing the Best-of-N baseline (53.4\%). On Game24, AlphaZero similarly degrades at Large budget (82.9\%), whereas ReSCALE continues to improve (85.3\%), both outperforming the Best-of-N baseline (54.1\%).}
    \label{table:results}
\end{table*}

\section{Experiments}\label{sec:experiments}

The goal of our experiments is to compare the performance of the standard AlphaZero tree search with its Gumbel AlphaZero variant.
We follow the experimental setup from prior work \citep{wan2024alphazero} where the AlphaZero tree search for LLM decoding was proposed.

\noindent In particular, we evaluate both methods on tasks that use sentence-level actions.
We consider the mathematical reasoning dataset GSM8K~\citep{cobbe2021training} and the mathematical planning task Game24~\citep{yao2023tree}.
We use Llama2-7B~\citep{touvron2023llama} as the base language model.
The \texttt{DELIM} token is the newline character.

\paragraph{Experimental Setup}
We first perform supervised fine-tuning (SFT) of Llama2-7B on each task's training set. 
For every example, the target is constructed by concatenating all chain-of-thought steps (separated by the \texttt{DELIM} token), followed by the final line: ``The answer is \{answer\}.'' 
SFT is performed for three epochs with end-of-epoch checkpoints.

We then train the value network $v_{\phi}$. 
To construct the value-training dataset, we sample 100 completions per example from the task's training set using each SFT checkpoint with temperature 0.7, remove duplicates, and retain 17 unique completions per checkpoint (51 in total). 
The completions are split into sentence-level actions using the \texttt{DELIM} token. 
For each completion, we locate the first action containing ``The answer is \{answer\}'' and check whether the predicted answer is correct. 
If the answer is correct, that state receives a reward of 1; otherwise, it receives 0. All preceding states receive a reward of 0. 
Value targets are computed using a Monte Carlo estimate of the return. 
The value network shares the LLM backbone with an additional MLP head and is trained for three epochs.

\paragraph{Experimental Results}

We compare AlphaZero and ReSCALE tree search across three budget levels, progressively increasing tree width, depth, and the number of simulations. We group these hyperparameters into three levels, called Small, Medium, and Large, based on their token usage. As an extra baseline, we evaluate Best-of-N by generating candidates with the SFT Llama2-7B at temperature 1.0 and scoring them with the value network $v_{\phi}$. Results are reported in Table~\ref{table:results}.

On GSM8K and Game24, ReSCALE exhibits a steady increase in accuracy as the compute budget grows. In contrast, AlphaZero's accuracy plateaus and even degrades as the budget increases from Medium to Large on GSM8K and at Large budget on Game24.

\begin{takeaway}
\textbf{\textit{Takeaway:}} We analyzed MCTS trees for examples where AlphaZero MCTS produced incorrect answers and found that it overexploits a small subset of actions, leading to a large imbalance in visit counts between explored and under-explored nodes. This issue worsens with larger budgets, as root selection favors the most visited node and ignores potentially better alternatives. ReSCALE mitigates this by using Sequential Halving to preserve exploration diversity at each step.
\end{takeaway}

\paragraph{Ablation Studies}
We perform ablations to isolate the effects of Gumbel noise and Sequential Halving. 
To remove Gumbel noise, we set it to zero. 
To remove Sequential Halving, we replace it at the root node with the PUCT rule from AlphaZero MCTS while still adding Gumbel noise to the root prior.

Experiments are conducted on GSM8K using three seeds, with 50 simulations, a maximum width of 24, and a maximum depth of 16.
Our method achieves $60.1 \pm 0.9\%$ accuracy. 
Removing Sequential Halving reduces accuracy by 4.7\%, while removing Gumbel noise reduces it by 1.7\%.
These results highlight the importance of both Gumbel noise and Sequential Halving for the performance of ReSCALE.

\section{Conclusion}

We presented ReSCALE, which replaces PUCT selection and Dirichlet noise with Sequential Halving and Gumbel sampling for LLM reasoning.
On GSM8K and Game24, it restores monotonic scaling, reaching 58.4\% and 85.3\%, respectively, where the baseline degrades to 53.6\% and 82.9\%.
Ablations confirm Sequential Halving as the primary contributor.
These results show that root action-selection design is critical for tree search in LLM decoding, and that fixed-budget best-arm identification techniques can address scaling pathologies.

\newpage
\bibliography{aaai2026}

\end{document}